\documentclass[conference]{IEEEtran}
\IEEEoverridecommandlockouts

\usepackage{cite}
\usepackage{amsmath,amssymb,amsfonts}
\usepackage{algorithmic}
\usepackage{graphicx}
\usepackage{textcomp}
\usepackage{xcolor}
\usepackage{booktabs}
\usepackage{multirow}
\usepackage{makecell}
\usepackage{array}
\usepackage{siunitx}

\newcommand{\tablesize}{\footnotesize}
\renewcommand{\arraystretch}{1.15}
\newcolumntype{L}[1]{>{\raggedright\arraybackslash}p{#1}}

\begin{document}

\title{Classifier-Centric Adaptive Framework for Open-Vocabulary Camouflaged Object Segmentation}

\author{
\IEEEauthorblockN{Hanyu Zhang\textsuperscript{*}}
\IEEEauthorblockA{\textit{School of Automation} \\
\textit{Southeast University}\\
Nanjing, China \\
213233503@seu.edu.cn}
\and
\IEEEauthorblockN{Yiming Zhou\textsuperscript{*}}
\IEEEauthorblockA{\textit{School of Automation} \\
\textit{Southeast University}\\
Nanjing, China \\
213230124@seu.edu.cn}
\and
\IEEEauthorblockN{Jinxia Zhang\textsuperscript{\dag}}
\IEEEauthorblockA{\textit{School of Automation} \\
\textit{Southeast University}\\
Nanjing, China \\
jinxiazhang@seu.edu.cn}
\thanks{\textsuperscript{*}\;Contribute equally to this work.\; \textsuperscript{\dag}\;Corresponding author.}
}

\maketitle

\begin{abstract}
Open-vocabulary camouflaged object segmentation requires models to segment camouflaged objects of arbitrary categories unseen during training, placing extremely high demands on generalization capabilities. Through analysis of existing methods, it is observed that the classification component significantly affects overall segmentation performance. Accordingly, a classifier-centric adaptive framework is proposed to enhance segmentation performance by improving the classification component via a lightweight text adapter with a novel layered asymmetric initialization. Through the classification enhancement, the proposed method achieves substantial improvements in segmentation metrics compared to the OVCoser baseline on the OVCamo benchmark: cIoU increases from 0.443 to 0.493, cSm from 0.579 to 0.658, and cMAE reduces from 0.336 to 0.239. These results demonstrate that targeted classification enhancement provides an effective approach for advancing camouflaged object segmentation performance.
\end{abstract}

\begin{IEEEkeywords}
Open-vocabulary camouflaged object segmentation, text adapter, initialization strategy, vision-language model, parameter-efficient fine-tuning
\end{IEEEkeywords}

\section{Introduction}
Open-vocabulary camouflaged object segmentation (OVCOS) is a challenging task that requires segmenting camouflaged objects from unseen categories~\cite{b1}. The difficulty arises from combining visual ambiguity in camouflaged scenes with the need for semantic generalization to novel object classes.

While existing methods focus on closed-vocabulary scenarios~\cite{b16,b17}, recent work like OVCoser~\cite{b1} has shown the feasibility of open-vocabulary settings. Experimental analysis reveals that improving classification accuracy leads to substantial segmentation performance gains, indicating that enhanced semantic understanding is key to advancing OVCOS.

Based on this insight, this work proposes a classifier-centric framework that improves OVCOS through lightweight text adapters with a novel layered asymmetric initialization strategy. The approach achieves significant improvements on the OVCamo benchmark across all metrics.

The main contributions include: \textbf{(i)} demonstrating that classification enhancement directly improves OVCOS performance; \textbf{(ii)} proposing an effective layered asymmetric initialization for lightweight adapters; \textbf{(iii)} achieving state-of-the-art results on OVCamo benchmark.
\section{Method}

\subsection{Method Overview}
The proposed framework improves OVCOS performance by enhancing classification capabilities through iterative mask-guided refinement. The approach employs a two-stage inference process that leverages the synergy between segmentation and classification components.

The framework consists of three key components: (1) an enhanced Alpha-CLIP~\cite{b3} classifier with lightweight text adapters, (2) a layered asymmetric initialization strategy for improved adapter performance, and (3) an iterative refinement process where predicted masks guide classification improvement.

\begin{figure*}[t]
\centering
\includegraphics[width=0.85\textwidth]{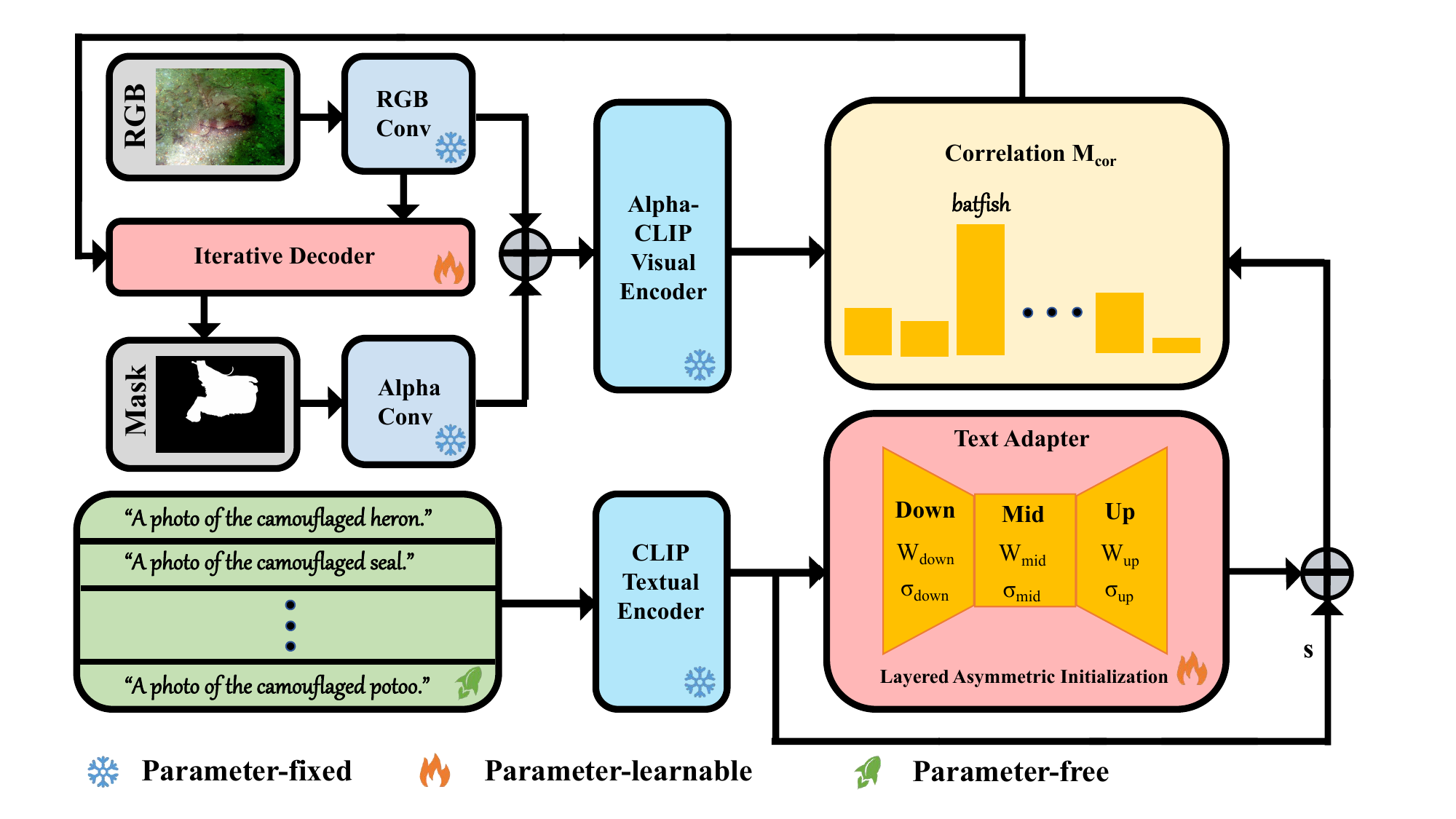}
\caption{Overall architecture of the proposed framework. RGB images and masks are processed through separate pathways in Alpha-CLIP. A lightweight text adapter with layered asymmetric initialization enhances the textual encoder for improved classification.}
\label{fig:overall_architecture}
\end{figure*}

As illustrated in Figure~\ref{fig:overall_architecture}, the architecture consists of two main processing pathways. The visual pathway processes RGB input images and predicted masks through Alpha-CLIP's visual encoder to extract multimodal features. On the textual side, class-specific prompts(``A photo of the camouflaged $\langle \text{class}\rangle$''
)
 are processed by the frozen CLIP textual encoder~\cite{b2} enhanced with a lightweight text adapter at the final layer.

The text adapter employs a three-layer bottleneck architecture with projection matrices $W_{\text{down}}$, $W_{\text{mid}}$, and $W_{\text{up}}$, initialized using the proposed layered asymmetric initialization where $\sigma_{\text{down}} > \sigma_{\text{mid}} > \sigma_{\text{up}}$. The adapter output is scaled by a learnable factor $s$ and added via residual connection.

The iterative refinement process generates initial mask predictions, which are fed back to provide spatial guidance for improved classification. This creates a feedback loop where better masks enable more accurate classification. The framework uses frozen CLIP backbone networks to preserve pre-trained knowledge while fine-tuning only the text adapter for efficient adaptation.

\subsection{Lightweight Text Adapter}
To achieve efficient domain-specific semantic enhancement for camouflaged object recognition, a lightweight text adapter is designed and inserted at the final layer of the text encoder. Let $\mathbf{z}$ denote the input text feature at this layer. The adapter transformation is defined as:
\begin{equation}
A^{t}(\mathbf{z}) = W^{t}_{\mathrm{up}} \cdot \delta\!\left( W^{t}_{\mathrm{mid}} \cdot \delta\!\left( W^{t}_{\mathrm{down}} \cdot \mathbf{z} \right)\right),
\end{equation}
where $W^{t}_{\mathrm{down}} \in \mathbb{R}^{r \times d}$ projects features to a bottleneck with $r \ll d$, $W^{t}_{\mathrm{mid}} \in \mathbb{R}^{r \times r}$ applies nonlinear transformation, $\delta(\cdot)$ denotes ReLU activation~\cite{b5}, and $W^{t}_{\mathrm{up}} \in \mathbb{R}^{d \times r}$ restores the original dimension.

The residual connection with learnable scaling factor $s$ is applied as:
\begin{equation}
\label{eq:residual_update}
\mathbf{y} = \mathbf{z} + s \cdot A^{t}(\mathbf{z}).
\end{equation}
This bottleneck design enables nonlinear refinement of camouflage-related semantics while maintaining low parameter overhead.

The system processes RGB images and masks through separate pathways in Alpha-CLIP's visual encoder. Class labels are embedded into prompt templates ("A photo of the camouflaged $\langle\text{class}\rangle$") and processed by the CLIP textual encoder with our adapter at the final layer. The [EOS] token is used for text-image similarity computation following standard protocols.

\subsection{Layered Asymmetric Initialization}
Through extensive empirical exploration, it is discovered that lightweight adapter architectures exhibit high sensitivity to initialization strategies due to their compact parameter space. Based on this observation, a layered asymmetric initialization (LAI) strategy is proposed that assigns different variances to different projection layers to achieve better convergence and performance. This initialization approach is designed to be applicable to similar lightweight adapter architectures beyond the current framework.

The proposed LAI strategy assigns different initialization variances to the three projections while keeping the residual gate small at the beginning of training. The adapter-induced residual update follows Eq.~\eqref{eq:residual_update}, and the projection matrices are initialized with zero-mean Gaussians under:
\begin{align}
W^{t}_{\mathrm{down}} &\sim \mathcal{N}\!\big(0,\,\sigma_{\mathrm{down}}^{2}\big),\\
W^{t}_{\mathrm{up}}   &\sim \mathcal{N}\!\big(0,\,\sigma_{\mathrm{up}}^{2}\big),\\
W_{\mathrm{mid}}      &\sim \mathcal{N}\!\big(0,\,\sigma_{\mathrm{mid}}^{2}\big),
\end{align}
with the ordering:
\begin{equation}
\sigma_{\mathrm{down}} > \sigma_{\mathrm{mid}} > \sigma_{\mathrm{up}}.
\end{equation}
The residual scale $s$ is learnable and is initialized with a small value $s_0$ to ensure stable training at the beginning:
\begin{equation}
s_{0} \in [\alpha_{\min}, \alpha_{\max}],
\end{equation}
after which $s$ is optimized jointly with the adapter.

This asymmetric ordering is designed to create an effective balance between exploration capacity and initialization stability. The larger variance in $W_{\mathrm{down}}$ enables broader feature space exploration in the compressed representation, facilitating the discovery of camouflage-specific semantic patterns. The smaller variance in $W_{\mathrm{up}}$ preserves the stability of the pre-trained representation structure, ensuring that beneficial knowledge from pre-training is maintained. The middle transformation $W_{\mathrm{mid}}$ provides nonlinear refinement at a moderate scale, balancing between feature transformation capability and training stability. This ordering facilitates efficient adaptation while maintaining beneficial inductive biases from pre-training.

The pronounced impact of initialization in our framework can be attributed to the lightweight nature of our adapter. With a compact parameter space of approximately 0.18M parameters compared to the frozen CLIP backbone, each parameter carries relatively more influence on the final model behavior. This parameter efficiency amplifies the importance of proper initialization, as small variations in initial weights can lead to substantially different optimization trajectories and convergence points.

The initial perturbation magnitude is controlled by the product of all layer norms and the scaling factor $s$. The ordering $\sigma_{\mathrm{down}} > \sigma_{\mathrm{mid}} > \sigma_{\mathrm{up}}$ ensures that the adapter starts with moderate perturbations to the pre-trained features while allowing sufficient parameter diversity for effective adaptation. The small initial value of $s$ provides additional stability control during early training stages.
\subsection{Test-Time Augmentation}
To further enhance classification robustness, a test-time augmentation (TTA) strategy is employed. Multiple scale factors and horizontal flips are applied to generate diverse views of the input image. The final prediction aggregates results from all augmented views using confidence-weighted voting:
\begin{equation}
P_{\text{final}}(c|I) = \frac{\sum_{i=1}^{N} w_i \cdot \text{softmax}(\mathbf{z}_i / \tau)}{\sum_{i=1}^{N} w_i}
\end{equation}
where $w_i = \max(\text{softmax}(\mathbf{z}_i))$ represents the confidence weight for each view and $\tau$ is the temperature parameter.

\section{Experimental Results and Analysis}

\subsection{Experimental Setup}
\textbf{Implementation details.} The framework is implemented using PyTorch on NVIDIA RTX 4090 GPUs . The method is built upon ViT-L/14@336px backbone with Alpha-CLIP~\cite{b3} pre-trained weights. The text adapter employs bottleneck dimension $r=64$ with learnable scaling factor initialized to 0.15. SGD optimizer is used with learning rate $0.0035$, momentum $0.9$, and weight decay $1\times 10^{-5}$. Training is conducted for 10 epochs with batch size 16. Input images are resized to $336\times336$ pixels. Prompt template and class-text construction follow OVCOS~\cite{b1}. The construction of mask follows Alpha-CLIP~\cite{b3}. LAI initialization uses the specified variances with uniform sampling for the residual scale $s$ within the range $[0.075, 0.225]$ for reproducibility.

\textbf{Training details.} The approach requires training only the lightweight text adapter (approximately 0.18M parameters) while keeping the CLIP backbone frozen, significantly reducing computational requirements compared to full model fine-tuning. Training requires approximately 30 minutes on a single RTX 4090.

\subsection{Dataset and Evaluation Protocol}
Evaluation is conducted on the OVCamo benchmark for OVCOS~\cite{b1},which contains 11{,}483 images across 75 categories.The dataset split follows~\cite{b1} with 7{,}713 images from 14 seen categories for training and 3{,}770 images from 61 unseen categories for testing.

For classification evaluation, top-1 classification accuracy is reported on correctly identified camouflaged regions. For segmentation evaluation, standard metrics including cIoU (class-aware Intersection over Union), cSm (class-aware S-measure), cMAE (class-aware Mean Absolute Error), cF$_\beta$ (class-aware F-measure), cF$_{\omega\beta}$ (class-aware weighted F-measure), and cEm (class-aware E-measure) are employed to comprehensively assess segmentation performance.

\subsection{Classification Performance Comparison}
This section evaluates the classification performance of different methods under various inference scenarios to demonstrate the effectiveness of our approach. The results are presented in Table~\ref{tab:classification_results}.

\begin{table}[t]
\centering
\tablesize
\caption{Quantitative comparison of classification results with existing methods on OVCamo dataset. Numbers show absolute Top-1 accuracy (\%) with $\Delta$ indicating gain over Alpha-CLIP baseline (in pp).}
\label{tab:classification_results}
\begin{tabular}{lcc}
\toprule
\textbf{Method} & \textbf{Truth (GT mask)} & \textbf{All-black} \\
\midrule
Alpha-CLIP (2024)~\cite{b3} & 74.60\% & 69.88\% \\
CLIP-Adapter (2021)~\cite{b18} & 74.06\% & 71.06\% \\
CoT-LLM w/ CoVP (2024)~\cite{b19} & 75.40\% & 75.40\% \\
\midrule
Standard Adapter (Ours) & 74.83\% & 73.51\% \\
Standard Adapter + LAI (Ours) & 78.95\% & 75.87\% \\
\textbf{Standard Adapter + LAI + TTA (Ours)} & \textbf{79.75\%} & \textbf{76.85\%} \\
\bottomrule
\end{tabular}
\end{table}

The proposed method achieves significant improvements in classification accuracy across both inference scenarios. With ground-truth masks, our complete method (LAI + TTA) reaches 79.75\%, representing a 5.15 percentage point improvement over the Alpha-CLIP baseline. In the more challenging all-black scenario without spatial guidance, the improvement is even more substantial at 6.97 percentage points (from 69.88\% to 76.85\%).

The LAI adapter alone provides the majority of performance gains, demonstrating its effectiveness in enhancing semantic understanding for camouflaged objects. TTA contributes additional consistent improvements of approximately 0.8-1.0 percentage points across both scenarios.

\subsection{Segmentation Performance Comparison}
This section compares segmentation performance across different methods to validate that improved classification translates to better segmentation results. Table~\ref{tab:ovcos_s3} shows the comparison, and Table~\ref{tab:classification_impact} demonstrates the impact of classification enhancement on segmentation metrics.

\begin{table}[t]
\centering
\tablesize
\caption{Quantitative comparison of segmentation results with existing methods on OVCamo dataset. Metrics: cSm$\uparrow$/cF$\omega\beta$$\uparrow$/cMAE$\downarrow$/cF$\beta$$\uparrow$/cEm$\uparrow$/cIoU$\uparrow$.}
\label{tab:ovcos_s3}
\begingroup
\setlength{\tabcolsep}{3.8pt}
\renewcommand{\arraystretch}{1.12}
\resizebox{\columnwidth}{!}{%
\begin{tabular}{lcccccc}
\toprule
\textbf{Method} & \textbf{cSm} & \textbf{cF$\omega\beta$} & \textbf{cMAE} & \textbf{cF$\beta$} & \textbf{cEm} & \textbf{cIoU} \\
\midrule
SimSeg'21~\cite{b6}      & 0.053 & 0.049 & 0.921 & 0.056 & 0.098 & 0.047 \\
OVSeg'22~\cite{b7}       & 0.024 & 0.046 & 0.954 & 0.056 & 0.130 & 0.046 \\
ODISE'23~\cite{b10}      & 0.187 & 0.119 & 0.700 & 0.211 & 0.298 & 0.167 \\
SAN'23~\cite{b9}         & 0.275 & 0.202 & 0.612 & 0.220 & 0.318 & 0.189 \\
CAT-Seg'23~\cite{b11}    & 0.181 & 0.106 & 0.719 & 0.123 & 0.196 & 0.094 \\
FC-CLIP'23~\cite{b12}    & 0.080 & 0.076 & 0.872 & 0.090 & 0.191 & 0.072 \\
OVCoser'24~\cite{b1} & 0.579 & 0.490 & 0.336 & 0.520 & 0.616 & 0.443 \\
\midrule
\textbf{Ours} & \textbf{0.658} & \textbf{0.547} & \textbf{0.239} & \textbf{0.582} & \textbf{0.696} & \textbf{0.493} \\
\bottomrule
\end{tabular}}
\endgroup
\end{table}

The proposed method achieves substantial improvements across all segmentation metrics compared to existing approaches. The cIoU increases from 0.443 to 0.493, cSm from 0.579 to 0.658, and cMAE reduces from 0.336 to 0.239 when compared to the OVCoser baseline. These improvements demonstrate that enhanced classification capabilities directly translate to better segmentation performance.

Compared to general open-vocabulary segmentation methods, the specialized approach shows significant advantages in handling camouflaged objects, highlighting the importance of domain-specific adaptations for this challenging task.

\subsection{Ablation Study}

\subsubsection{Impact of Classification Enhancement on Segmentation Performance}
The classification experiment in Table~\ref{tab:classification_impact} provides strong evidence for the proposed approach. When perfect classification is achieved, segmentation performance improves dramatically across all metrics, with cMAE dropping from 0.336 to 0.036 and notable gains in other measures. This validates the hypothesis that improving classification accuracy leads to meaningful overall performance improvements in open-vocabulary camouflaged object segmentation.

\begin{table}[t]
\centering
\tablesize
\caption{Ablation study on the effect of classification enhancement on segmentation performance.}
\label{tab:classification_impact}
\begingroup
\setlength{\tabcolsep}{3.5pt}
\renewcommand{\arraystretch}{1.08}
\resizebox{\columnwidth}{!}{%
\begin{tabular}{lcccccc}
\toprule
\textbf{Setting} & \textbf{cSm} & \textbf{cF$\omega\beta$} & \textbf{cMAE} & \textbf{cF$\beta$} & \textbf{cEm} & \textbf{cIoU} \\
\midrule
CLIP Classifier & 0.584 & 0.513 & 0.336 & 0.538 & 0.616 & 0.447 \\
\midrule
\textbf{Ours (LAI + TTA)} & \textbf{0.658} & \textbf{0.547} & \textbf{0.239} & \textbf{0.582} & \textbf{0.696} & \textbf{0.493} \\
\textbf{Oracle Classifier} & \textbf{0.800} & \textbf{0.628} & \textbf{0.036} & \textbf{0.672} & \textbf{0.856} & \textbf{0.570} \\
\bottomrule
\end{tabular}}
\endgroup
\end{table}

\subsubsection{Effectiveness of Different Components}
This section analyzes the contribution of each component in the framework through systematic ablation experiments. The results are shown in Table~\ref{tab:ablation}.

\begin{table}[t]
\centering
\tablesize
\caption{Ablation study on OVCamo. Numbers show absolute Top-1 accuracy (\%) with $\Delta$ indicating gain over Alpha-CLIP baseline (in pp).}
\label{tab:ablation}
\begin{tabular}{lcc}
\toprule
\textbf{Components} & \textbf{Truth (GT mask)} & \textbf{All-black} \\
\midrule
Alpha-CLIP (baseline)~\cite{b3} & 74.60\% & 69.88\% \\
+ Standard Adapter & 74.83\% \;($+0.23$) & 73.51\% \;($+3.63$) \\
+ LAI & 78.95\% \;($+4.35$) & 75.87\% \;($+5.99$) \\
+ TTA & \textbf{79.75\%} \;($\mathbf{+5.15}$) & \textbf{76.85\%} \;($\mathbf{+6.97}$) \\
\bottomrule
\end{tabular}
\end{table}

The ablation results demonstrate the effectiveness of each proposed component. The layered asymmetric initialization (LAI) provides the most significant contribution, achieving 4.35 percentage point improvement in the Truth setting and 5.99 points in the All-black setting compared to the Alpha-CLIP baseline.

The superiority of LAI over standard adapter initialization is evident: LAI achieves 78.95\% vs 74.83\% (+4.12pp) in the Truth setting and 75.87\% vs 73.51\% (+2.36pp) in the All-black setting. This substantial performance difference highlights the importance of proper initialization strategies for lightweight adapters, where each parameter carries significant influence due to the compact parameter space.

Test-time augmentation consistently provides additional improvements of approximately 0.8-1.0 percentage points across both scenarios, demonstrating its effectiveness as a complementary enhancement technique.

The segmentation-guided inference approach (using predicted masks from segmentation models) achieves 79\% accuracy, representing a practical middle ground between the all-black scenario (76.85\%) and the ground-truth scenario (79.75\%). This demonstrates the value of iterative mask-guided refinement in real-world applications where ground-truth masks are unavailable.

\section{Conclusion}
It is discovered that improving classification accuracy can lead to meaningful improvements in open-vocabulary camouflaged object segmentation through a classifier-centric adaptive framework. The proposed lightweight text adapters with empirically effective layered asymmetric initialization and iterative mask-guided refinement enable efficient semantic enhancement while preserving pre-trained knowledge. On the OVCamo benchmark, substantial improvements in segmentation metrics are achieved compared to existing methods: cIoU increases from 0.443 to 0.493, cSm from 0.579 to 0.658, and cMAE reduces from 0.336 to 0.239, demonstrating that targeted classification enhancement provides an effective approach for advancing OVCOS capabilities. The significant impact of the initialization strategy underscores the importance of proper parameter initialization in lightweight adapter architectures.

While the LAI strategy shows consistent empirical benefits, future work could explore the theoretical foundations of these initialization strategies and their connection to adapter optimization dynamics in vision-language models. Additionally, investigating the integration of this framework with other advanced segmentation architectures and extending the approach to other challenging vision tasks could further validate its generalizability.

\end{document}